\documentclass[letterpaper, 10 pt, conference]{ieeeconf}  % Comment this line out if you need a4paper
\IEEEoverridecommandlockouts                              % This command is only needed if
\usepackage{graphics}    % for pdf, bitmapped graphics files
\usepackage{amsmath}     % assumes amsmath package installed
\usepackage{amssymb}     % assumes amsmath package installed
\usepackage{graphicx}
\usepackage[separate-uncertainty=true]{siunitx}%
\usepackage[caption=false, font=footnotesize]{subfig}
\usepackage{tabularx}
\def\tabref#1{Tab.~\ref{#1}}
\def\eqref#1{Eq.~(\ref{#1})}

\newcommand\etal{\emph{et al. }}

\title{\LARGE \bf Learning Personalized Human-Aware Robot Navigation\\Using Virtual Reality Demonstrations from a User Study}

\author{Jorge de Heuvel$^1$ \and Nathan Corral$^1$ \and Lilli Bruckschen$^2$ \and Maren Bennewitz$^1$% <-this % stops a space
  \thanks{$^1$ University of Bonn, Germany.}%
  \thanks{$^2$ Fraunhofer FKIE, Bonn, Germany.}%
  \thanks{This work has partially been funded by the Deutsche Forschungsgemeinschaft (DFG, German
  		Research Foundation) under the grant number BE~4420/2-2~(FOR 2535 
Anticipating Human Behavior).}
}

\begin{document}
\maketitle
\thispagestyle{empty} 
\pagestyle{empty}

%%%%%%%%%%%%%%%%%%%%%%%%%%%%%%%%%%%%%%%%%%%%%%%%%%%%%%%%%%%%%%%%%%%%%%%%%%%%%%%%
\begin{abstract} 
  % ´Personalized robot behavior is the next level.
  % WHY is it relevant?
  % WHICH PROBLEM do we address?
  % HOW is our approach special, WHAT are we actually doing, and WHAT IS NEW
  %% IMPLEMENTATION, EVALUATION, WHAT FOLLOWS 
  For the most comfortable, human-aware robot navigation, subjective user preferences need to be taken into account.
  This paper presents a novel reinforcement learning framework to train a personalized navigation controller along with an intuitive virtual reality demonstration interface. 
  The conducted user study provides evidence that our personalized approach significantly outperforms classical approaches with more comfortable human-robot experiences. 
  We achieve these results using only a few demonstration trajectories from non-expert users, who predominantly appreciate the intuitive demonstration setup. As we show in the experiments, the learned controller generalizes well to states not covered in the demonstration data, while still reflecting user preferences during navigation. Finally, we transfer the navigation controller without loss in performance to a real robot.
\end{abstract}

%%%%%%%%%%%%%%%%%%%%%%%%%%%%%%%%%%%%%%%%%%%%%%%%%%%%%%%%%%%%%%%%%%%%%%%%%%%%%%%%
\section{Introduction}
\label{sec:intro}

%% WHY 
Robot personalization to specific user-preferences will become increasingly important, as robots find their way into our everyday life. Harmonic human-robot interactions build trust and satisfaction with the user \cite{gasteiger_factors_2021}, whereas negative interaction experiences can quickly lead to frustration \cite{kruse_human-aware_2013}. 
A cause for negative user experiences can be algorithms that do not reflect personal preferences.

\begin{figure} 
	\centering
	\subfloat{%
		\includegraphics[width=0.95\linewidth]{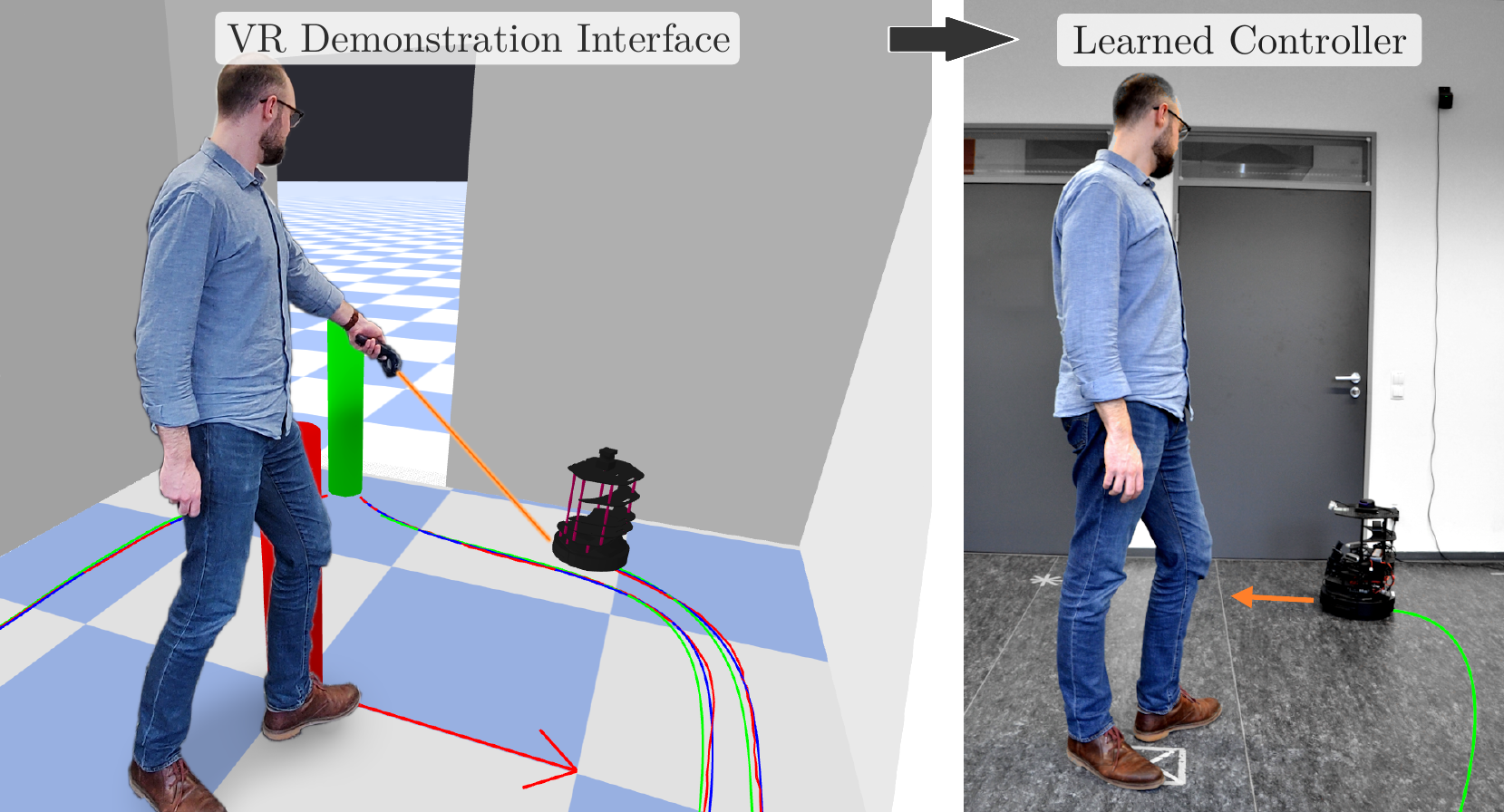}}
	\\
	\subfloat{%
		\includegraphics[width=0.95\linewidth]{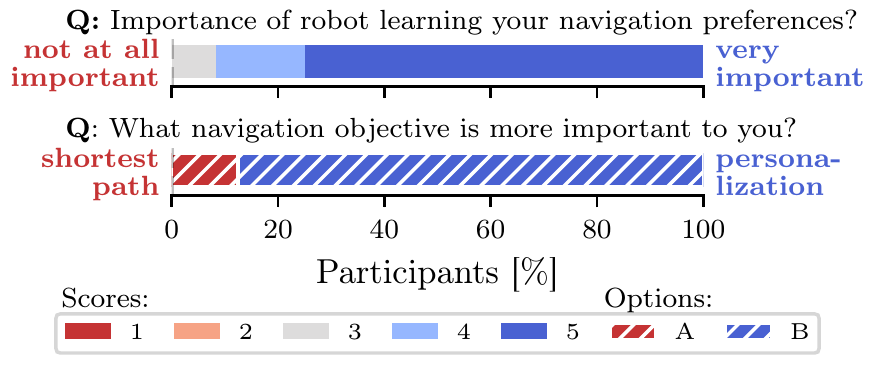}}
	\caption{
		\textbf{Top:} We propose a virtual reality (VR) interface to intuitively demonstrate robot navigation preferences by drawing trajectories onto the floor with a handheld controller. 
		\textbf{Bottom:} User study survey results on the importance of personalized navigation behavior. Participants strongly expressed their preference for personalization of robot navigation behavior, even at the possible cost of longer trajectories.
		\label{fig:motivation}}
\end{figure}

%% WHICH PROBLEM
Where mobile household robots navigate in the vicinity of a human, basic obstacle avoidance approaches fail to capture individual user preferences. 
While collision avoidance is undoubtedly crucial during navigation, the navigation policy should furthermore be human-aware and take into account user preferences regarding proxemics \cite{kruse_human-aware_2013} and privacy, compare Fig.~\ref{fig:motivation} (bottom). 
Subjective preferences may vary depending on the environment and social context, e.g., navigation preferences could reflect in the robot's approaching behavior, or always driving in front or behind the human.
In addition following a certain speed profile and maintaining a certain distance from humans and other obstacles in the environment might play a role.
The resulting navigation objective for the robot is to reach the navigation goal, not necessarily by only following the shortest path, but also by taking personal robot navigation preferences into account. 

Recent advances in learning socially-aware navigation behavior from human demonstrations have been made with inverse reinforcement learning, where the parameters of a proxemics-encoding reward function were inferred \cite{kollmitz_learning_2020}. Influenced by the initial shaping of the reward function \cite{ng_policy_1999}, such approaches lack the ability for navigation style personalization beyond the scope of the reward function. 
For smooth navigation, reinforcement learning (RL) based continuous control has lead to promising results on mobile robots \cite{tai_virtual--real_2017, pfeiffer_reinforced_2018}. Furthermore, off-policy RL methods can be complemented with demonstration data to greatly improve learning speed on a given task, even outperforming the resourcefulness of the original demonstrations \cite{vecerik_leveraging_2018}. However, RL robot navigation policies learn most efficient trajectories to the goal. These trajectories do not necessarily reflect the original demonstration behavior, which contains user preferences.
To more precisely imitate behavior from demonstrations, behavioral cloning (BC) can be used \cite{argall_survey_2009}. However, the final policy is limited by the quality and amount of demonstration data \cite{ravichandar_recent_2020}. The dataset would need to cover most of the state space to generalize fluently in unseen environments. This poses a problem, as human demonstrators can only provide limited amounts of demonstration data due to their finite patience~\cite{thomaz_reinforcement_nodate}.
%% HOW & WHAT
The question crystallizes, how do we efficiently record personal preferences and teach them to the robot, without being limited by the quality and quantity of demonstrations.

%\begin{figure} 
%	\centering
%	\subfloat{%
%		\includegraphics[width=0.95\linewidth]{figures/vr_demo_jorge}}
%	\\
%	\subfloat{%
%		\includegraphics[width=0.95\linewidth]{figures/motivation}}
%	\caption{
%		\textbf{Top:} We propose a virtual reality (VR) interface to intuitively demonstrate robot navigation preferences by drawing trajectories onto the floor with a handheld controller.
%		\textbf{Bottom:} User study survey results on the importance of personalized navigation behavior. Participants strongly expressed their preference for personalization of robot navigation behavior, even at the possible cost of longer trajectories.
%		\label{fig:motivation}}
%\end{figure}

%\begin{figure}[t]
%	\centering
%	\includegraphics[width=1\linewidth]{figures/motivation}
%	\caption{User study survey results on the importance of personalized navigation behavior. Participants strongly expressed their preference for personalization of robot navigation behavior, even at the possible cost of longer trajectories.
%		\label{fig:motivation}}
%\end{figure}

In order to solve the aforementioned challenges, we propose a novel navigation learning approach together with a virtual reality (VR) interface to intuitively demonstrate robot navigation preferences by drawing trajectories onto the floor with a handheld controller, see Fig.~\ref{fig:motivation}. 
Importantly, the interface does not require expert-level knowledge on robotics, facilitating personalized navigation to a wide range of users. 
Our demonstration process is time-efficient, as only few demonstrations are required. 
The demonstrations are leveraged to successfully train a personalized human-aware navigation controller, by combining deep reinforcement learning and behavioral cloning. 
We show that our navigation policy closely reflects user preferences from only a few demonstrations. But at the same time, it generalizes to unseen states.
In an extensive user study, we evaluate the personalized navigation behavior against classical navigation approaches both in VR and on a real robot.
% To the best of our knowledge, we present the first study on personalized navigation based on deep reinforcement learning from demonstrations.

%% MAIN CONTRIBUTION & WHAT FOLLOWS FROM THAT
The threefold \textbf{main contributions} of our study are:
\begin{itemize}
	\item A VR demonstration interface for teaching navigation preferences to robots intuitively.
	\item Learning a user-personalized, context-based navigation policy based on the combination of RL and BC.
	\item An interactive user study recording user specific navigation preferences, evaluating both the presented interface and learned personalized navigation policies.
\end{itemize}

%%%%%%%%%%%%%%%%%%%%%%%%%%%%%%%%%%%%%%%%%%%%%%%%%%%%%%%%%%%%%%%%%%%%%%%%%%%%%%%%
\section{Related Work}
\label{sec:related}
% Robot Personalization
Extensive research has been done on both human-aware navigation \cite{moller_survey_2021} and on robot personalization \cite{gasteiger_factors_2021, hellou_personalization_2021}, but surprisingly, very few can be found at the intersection of both disciplines.

% SOCIAL NAVIGATION
% proxemics 

% cost-maps
Various studies adapt human-aware navigation behavior either by learning or inferring cost-maps \cite{bungert_human-aware_2021, perez-higueras_teaching_2018, kollmitz_learning_2020}. 
These cost-maps usually encode proxemics or environmental characteristics.
% CONTEXT BASED NAVIGATION: Room Type
To improve navigation in human-robot interaction based on context, Bruckschen \etal \cite{bruckschen_human-aware_2020} leveraged previously observed human-object interactions to predict human navigation goals, which in turn enables foresighted robot navigation and assistance. 
Other studies aimed to distinguish between different environment types as context in order to automatically adjust the robot's navigation behavior \cite{xiao_appld_2020, zender_human-and_2007}.
In our work, we consider different environment scenarios as context.

% CONTEXT BASED NAVIGATION: Orientation, posture
Luber \etal \cite{luber_socially-aware_2012} studied the angle of approach between two individuals to improve human-aware navigation. 
Recently, Narayanan \etal \cite{narayanan_proxemo_2020} leveraged the human gait posture as social context for foresighted robot navigation by predicting the human's navigation intent and emotion. 
To build upon the aforementioned findings, we as well take the human orientation into account.

% LEARNING FROM DEMONSTRATIONS
To learn personal navigation preferences in a human-robot collaboration scenario from demonstrations, \mbox{Kollmitz~\etal\cite{kollmitz_learning_2020}} learned the parameters of a navigation reward-function from physical human-robot interaction via inverse reinforcement learning. 
More specifically, the navigation reward-function was learned from a user pushing the robot away to a desired distance. A limitation of this approach is the state space represented by a 2D grid map of the environment, making the approach unsuitable for larger and unknown environments. To overcome this limitation, our state space is robot-centric and continuous, focusing on the vicinity to the human and obstacles.

Xiao \etal \cite{xiao_appld_2020} proposed using teleoperation demonstrations to learn context-based parameters of a conventional planner. Here, the reproduction of demonstration trajectories during navigation is limited by the capabilities of the conventional planner. To ensure a more distinct preference reproduction including certain trajectory profiles, we chose a deep learning-based controller.

% REINFORCEMENT LEARNING + IMITATION LEARNING
To efficiently train a deep learning based navigation controller for robot navigation via reinforcement learning, Pfeiffer \etal \cite{pfeiffer_reinforced_2018} utilized demonstration navigation data gathered from an expert planner algorithm. The demonstration data was used to pre-train the agent via imitation learning, followed by the reinforcement learning. In our work, we use a similar architecture for continuous control learning, but in contrast, we focus on human demonstrations of robot trajectories.

% INTERFACE: WHAT ELSE?
Virtual reality environments have been successfully deployed to simulate human-robot interactions \cite{bungert_human-aware_2021, liu_understanding_2017}, offering a tool for realistic demonstration and evaluation. 
As a result, we chose to develop a VR interface that interactively records the user-demonstrated trajectories of a robot. 
These demonstrated trajectories give the data required to learn user-specific robot navigation preferences. 
The VR interface enables a first-person experience of the navigating robot during demonstration, ensuring a realistic perception of proxemic aspects. 
In these regards, a clear benefit over, e.g., real world robot teleoperation is the easy separation of the demonstration and reevaluation experience in simulation, enabling interactive replay of scenarios.

\section{Problem Definition and Assumptions}
In this work, we consider a differential wheeled robot that has a local navigation goal and navigates in the vicinity of a single human. 
Our goal is to create a personalized robot navigation controller that adapts to user preferences by learning from demonstrations of robot trajectories that include a velocity profile.
Hereby, we focus on local human avoidance taking into account user-specific preference.
Both human and robot are interacting in the same room, which serves as context for the navigation behavior.
We assume that the positions and orientations of the human, the robot, and all obstacles are known.
All parameters above can play a role for the robot navigation preferences of the user and need to be reflected in the robot-centric state space.

%%%%%%%%%%%%%%%%%%%%%%%%%%%%%%%%%%%%%%%%%%%%%%%%%%%%%%%%%%%%%%%%%%%%%%%%%%%%%%%%
\section{Reinforcement Learning from Demonstrations}
\label{sec:rl}
\begin{figure*}[ht!]
	\centering
	\includegraphics[width=0.95\linewidth]{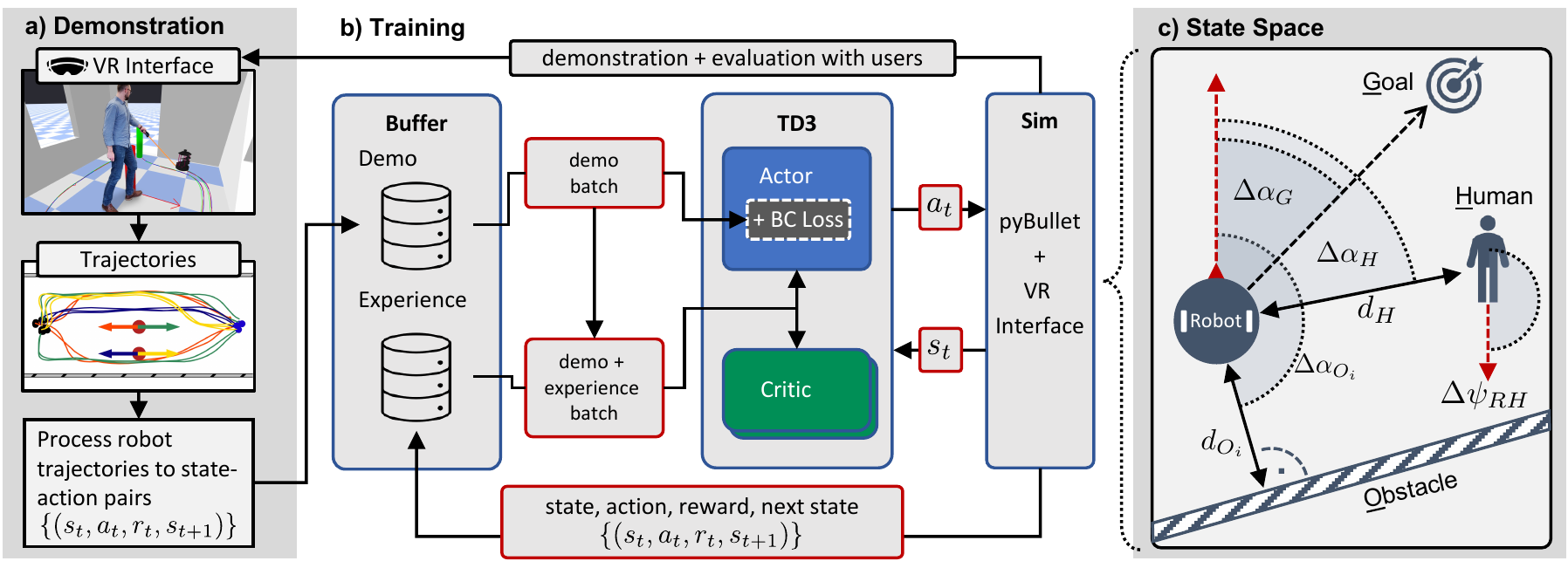}
	\caption{Schematic representation of the used architecture. \textbf{a)} Demonstration trajectories are drawn by the user and fed into the demonstration buffer. \textbf{b)} A TD3 reinforcement learning architecture with an additional behavioral cloning (BC) loss on the actor trains a personalized navigation policy for the human-robot interaction with continuous control. The learned policy is then evaluated in VR and subsequently transferred to a real robot. \textbf{c)} The robot-centric state space captures the vicinity and orientation of the human and the obstacles as well as the goal direction.}
	\label{fig:architecture}
\end{figure*}
We adapted a twin-delayed deep deterministic policy gradient (TD3) architecture consisting of an actor and two critic networks \cite{fujimoto_addressing_2018}. 
TD3 was chosen for two reasons: 
i) It has a continuous action space allowing smooth robot control and ii) it is off-policy, thus is a perfect candidate for use with demonstration data.
The actor network outputs two continuous robot control commands, i.e., forward and angular velocity. 
We introduce two modifications to classic TD3, similar to Nair \etal \cite{nair_overcoming_2018}: 
i) a behavioral cloning loss on the actor network and ii) a separate buffer to integrate demonstration data. 
The introduction of the behavioral cloning loss makes our approach a hybrid of reinforcement and imitation learning. 
Fig.~\ref{fig:architecture} depicts a schematic overview of our approach.

\subsection{Twin-Delayed Deep Deterministic Policy Gradient}
Reinforcement learning describes the optimization of transitions from state $s_t \rightarrow s_{t+1}$ following a Markov Decision Process that result in a reward $r_t = r(s_t, a_t)$, by taking an action $a_t = \pi_{\phi}(s_t)$  at time step $t$ with respect to a policy $\pi_{\phi}$. The tuples $\left(s_t, a_t, r_t, s_{t+1}\right)$ are referred to as state-action pairs. The optimization objective is to maximize the cumulative return $R = \sum^{T}_{i=t} \gamma^{(i-t)}r_t$ of the $\gamma$-discounted rewards, onward from $t$.
With TD3, we optimize the expected return 
\begin{align}
y_t = r_t + \gamma \min_{i=1, 2} Q_{\theta_i^*} \left( s_{t+1}, \pi_{\phi} \left( s_{t+1} \right) + \epsilon_{\theta_i} \right) \text{,}
\end{align}
while using the minimum of two critics $\left( Q_{\theta_1}, Q_{\theta_2}\right) $ to prevent value overestimation. $\theta_{i}$ denote the (network) parameters of critic $i$ and $\phi$ those of the actor. The clipped Gaussian policy noise $\epsilon_{\theta_i}$  stabilizes the Q-value estimation over similar state-action pairs and is controlled by the standard deviation $\sigma_{\epsilon_{\theta_i}}$

% Characteristic for TD3, actor and critic networks are optimized against a secondary so-called target network $\theta_{i}^{*}$ and $\phi^{*}$ respectively, which allows us to calculate the temporal difference. In comparison, those target networks are updated more slowly $\left( \theta_i^* \leftarrow \tau \theta_i + (1-\tau) \theta_i^* \text{, with } 0 < \tau < 1 \text{, analogous for }\phi \right) $ and provide a robust objective that stabilizes training.

To ensure sufficient exploration, we add Gaussian noise from a process $\mathcal{N}$ with standard deviation $\sigma_{\epsilon_\pi}$ to the actions drawn from the actor, so that $a_t = \pi_{\phi}(s_t) + \mathcal{N}(0, \sigma_{\epsilon_\pi})$.

% CRITIC LOSS 
To update the critic $\theta_{i}$, TD3 optimizes the loss
\begin{align}
\mathcal{L}_{\theta_i} = \frac{1}{b} \sum_{j}^{b} \left( y_j - Q_i \left(  s_j, a_j | \theta_{i} \right)  \right)^2
\end{align}
over all state-action pairs $j$ in the batch of size $b$. The actor network parameters $\phi_\pi$ are updated using the policy gradient:
% Actor Loss
\begin{align}
	\nabla_{\phi} J = \frac{1}{b} \sum_{j}^{b} \nabla_{a} Q_\text{min}(s, a|\theta) |_{s=s_j, a=\pi(s)} \nabla_{\phi} \pi(s|\phi)|_{s_j}
\end{align}
For further details on the learning algorithm, please refer to \cite{fujimoto_addressing_2018} and \cite{silver_deterministic_2014}.

%\subsection{Network Layout}
% NETWORKS
The actor and critic networks share a feed-forward three-layered perceptron architecture with 256 neurons each. We normalize both the input (observation space for actor and critic) and output (action space for actor) of the networks, respectively.

\subsection{Replay and Demonstration Buffer}
% Replay Buffer
In addition to TD3's standard experience replay buffer of size $B_{E}$, we introduce a second replay buffer to solely hold demonstration data, called the demonstration buffer. 
As the demonstration data is collected before training begins, its main difference to the experience buffer is that it is not updated during training and thus holds the demonstration data for the entire training duration. Its size $B_D$ is equivalent to the number of demonstration state-action pairs.

% Batch Size
We uniformly sample both from the experience replay buffer and the demonstration buffer with batch size $b_{E} = b_{D} = 64$. As both batches are merged, the actor and critic networks are optimized both with the demonstration and the latest experience data at every training step.

\subsection{Behavioral Cloning}
\label{sec:bc}
Similar to \cite{nair_overcoming_2018}, we introduce a behavioral cloning loss $\mathcal{L}_\text{BC}$ on the actor network as an auxiliary learning task:
% As a mean squared error loss, it represents the imitation learning part of the policy training:
\begin{align}
\mathcal{L}_\text{BC} = \sum_{i=1}^{b_D} || \pi(s_i|\phi) - a_i ||^2
\label{eq:bc_loss}
\end{align}
Only the batch fraction originating from the demonstration replay buffer is processed on the behavioral cloning loss.
The resulting gradient of the actor network is
\begin{align}
\nabla_{\phi} J_\text{total} = \lambda_{\text{RL}} \nabla_{\phi} J - \lambda_{\text{BC}}  \nabla_{\phi} \mathcal{L}_\text{BC} \text{.}
\end{align}
Leveling both gradients against each other using $\lambda_{\text{BC/RL}}$ is important to achieve a balance where the navigation policy reproduces demonstration-like behavior around known states~(in demonstration data), but also learns to handle unknown states correctly.

\subsection{State Space}
\label{sec:statespace}
A visualization of our robot-centric state space is shown in Fig.~\ref{fig:architecture}c. 
The state space is kept as minimalist as possible to ensure a fast and reliable training performance. 
%Motivated by the idea of robot personalization, we assume a single person in the navigation space of the robot. 
The functionality of our approach is proven for a single human in the vicinity of the robot.
%We assume the position of the human and the robot to be known.
The state~vector contains the person's distance $d_{H}$ to the robot's position and relative angle $\Delta\alpha_{H}$ to its orientation, facilitating human-awareness.
Furthermore, the relative angle to the navigation goal $\Delta \alpha_{G}$ is provided.
To increase awareness for the human's field of view, the person's body orientation relative to the orientation of the robot $\Delta \psi_{RH}$ is included. 
It indicates whether a person faces the robot or not. 
To deal with obstacles, we include the closest distance $d_{O_i}$ and relative angle $\Delta \alpha_{O_i}$ from the robot's pose to all environment obstacles $O_i$.

\begin{figure*}[t]
	\centering
	\includegraphics[width=0.95\linewidth]{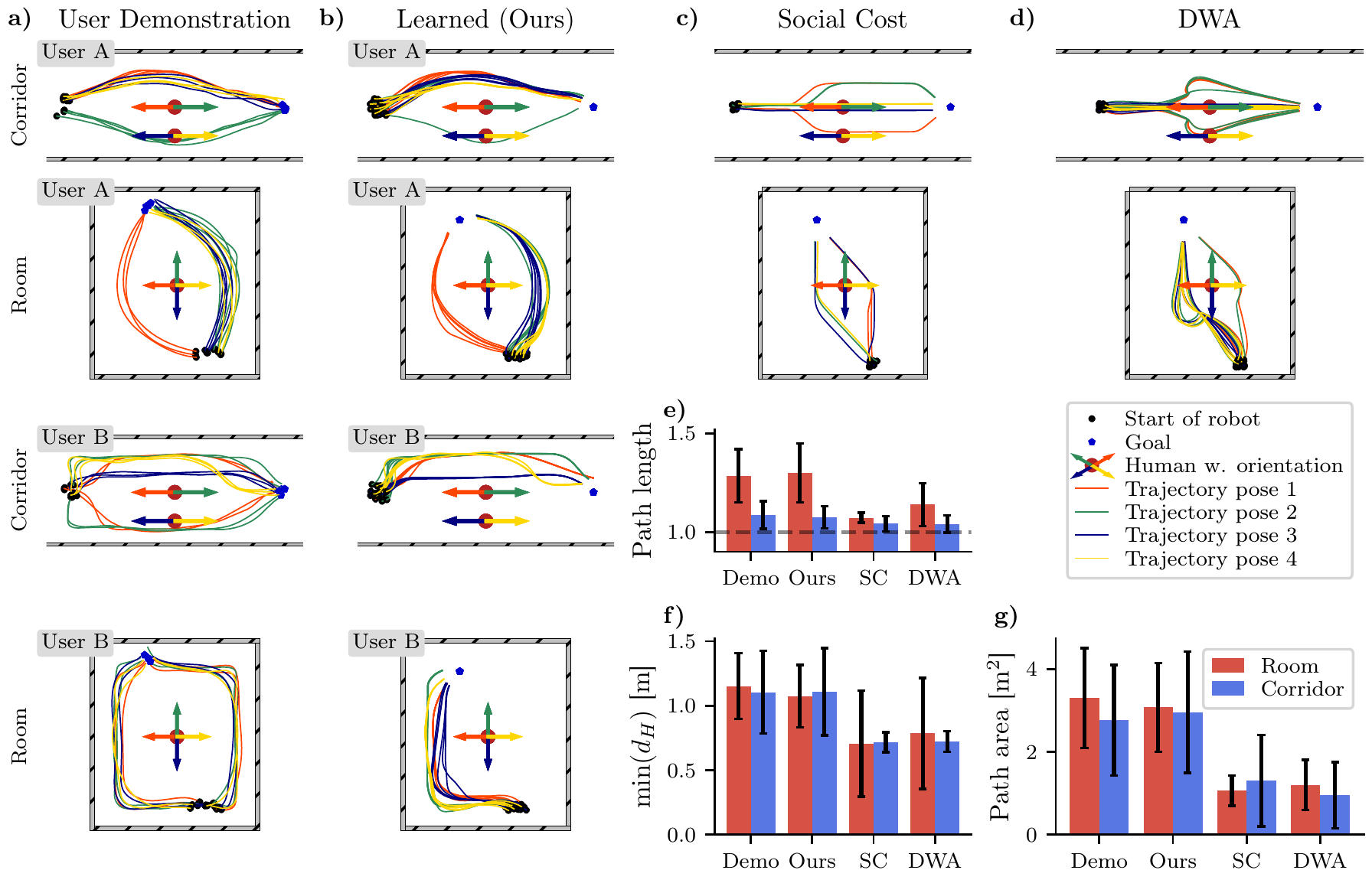}
	\caption{
		\textbf{a)} The demonstrated robot navigation preference trajectories of two participants A and B are shown for different human position-orientation pairs (color-coded). 
		Note the wall-following preference of user B, whereas user A prefers a smooth curve navigation style. 
		\textbf{b)}~The personalized controller successfully learned to reflect the individual user preferences. Note that when no specific side preference is given as in the demonstrations in the corridor, the controller reproduces trajectories mainly on one side.
		We evaluated our approach against \textbf{c)} the social cost model and \textbf{d)} the Dynamic Window Approach.
	%	The color shading of trajectories denotes the robot's forward velocity.
		A quantitative comparison of the different approaches in both environments reveals \textbf{e)}  a higher relative path length (normalized by linear distance) and \textbf{f)} a higher preferred minimum distance. 
		\textbf{g)} The increased path area for our controller (between the learned trajectory and linear distance) also points to a general preference for earlier deviation from the shortest path in favor for more comfortable trajectories.
	}
	\label{fig:approaches}
\end{figure*}

\subsection{Reward}
\label{sec:reward}
The reward function is designed to avoid collisions and ensure goal-oriented navigation behavior. We aim to teach user-specific navigation preferences not by complex reward shaping, but only via demonstration data. Consequently, we keep the reward as sparse as possible, besides basic collision penalties and goal rewards. More specifically, the reward function is defined as
\begin{align}
r = r_\text{collision} + r_\text{goal} + r_\text{timeout} \text{.}
\end{align}
We introduce a scaling factor for the reward $c_\text{rew} = 5$ that is used throughout the reward definition below. 
When the robot collides with the human or an obstacle during navigation, we penalize with  
\begin{align}
r_\text{collision} = 
\begin{cases} 
- c_\text{rew} & \text{if collision} \\
0 & \text{else.}
\end{cases}
\end{align}
The goal reaching reward is provided to the agent if the robot is located closer than a certain distance to the goal position:
\begin{align}
\label{eq:reward_goal}
r_\text{goal} = 
\begin{cases} 
+ c_\text{rew} & \text{if goal reached in demonstration data} \\
0 & \text{if goal reached during training} \\
0 & \text{else}
\end{cases}
\end{align}
Note that we give a detailed explanation on the goal reaching reward in Sec.~\ref{sec:demo_reward}. Finally,
the timeout reward encourages the agent to avoid inefficient actions by penalizing behavior where the goal is not reached by the agent after a certain number of steps~$N_\text{ep}$:
\begin{align}
r_\text{timeout} = 
\begin{cases} 
- \frac{c_\text{rew} }{2} & \text{if episode timeout } (n > N_\text{ep}) \\
0 & \text{else}
\end{cases}
\end{align}
All three conditions above (goal reached, collision, timeout) are end criteria for an episode.

\section{Demonstration and Training Environment}
\begin{figure}[ht!]
	\centering
	\subfloat{%
		\includegraphics[width=0.65\linewidth]{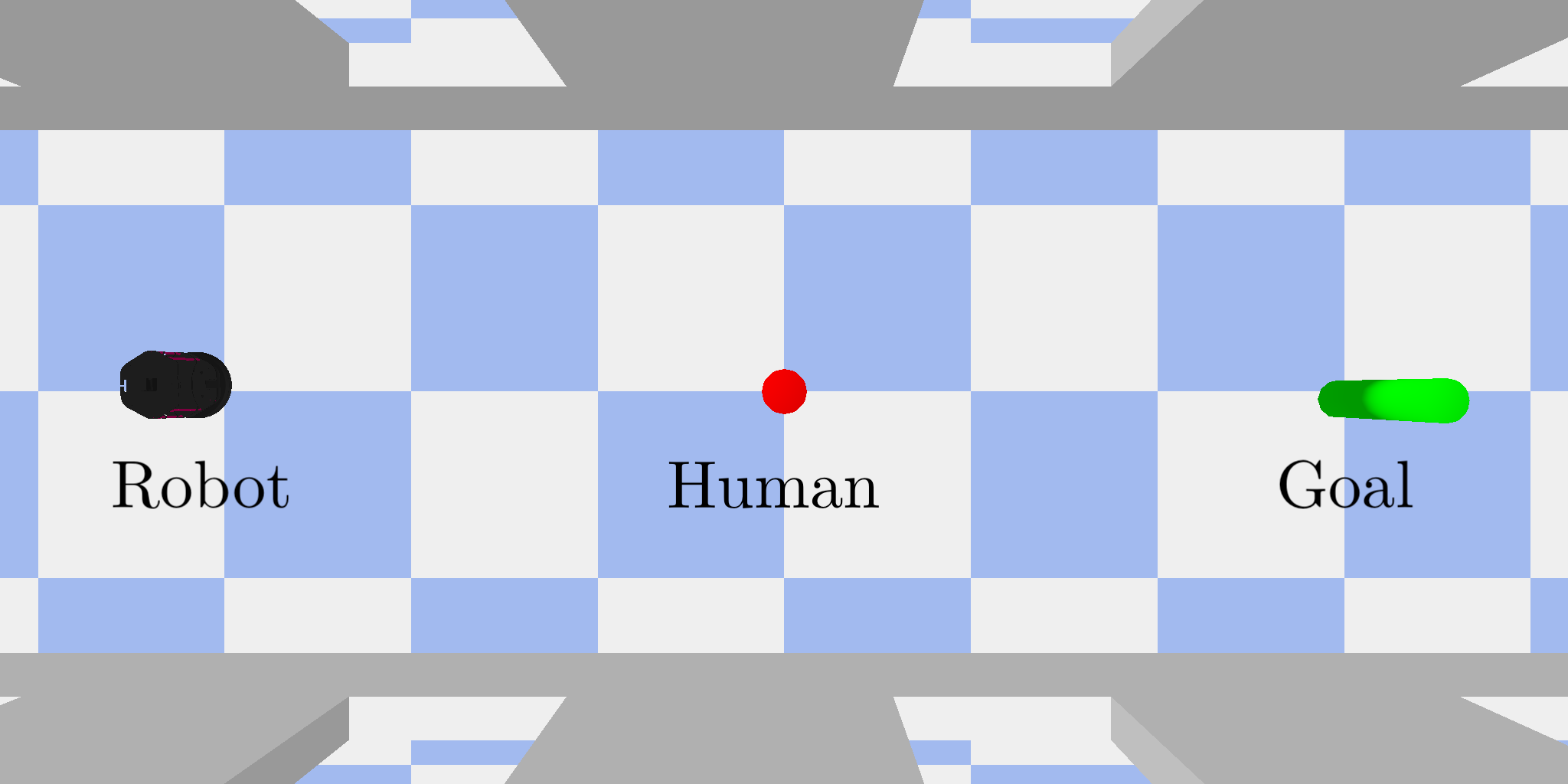}}
	\hfill
	\subfloat{%
		\includegraphics[width=0.325\linewidth]{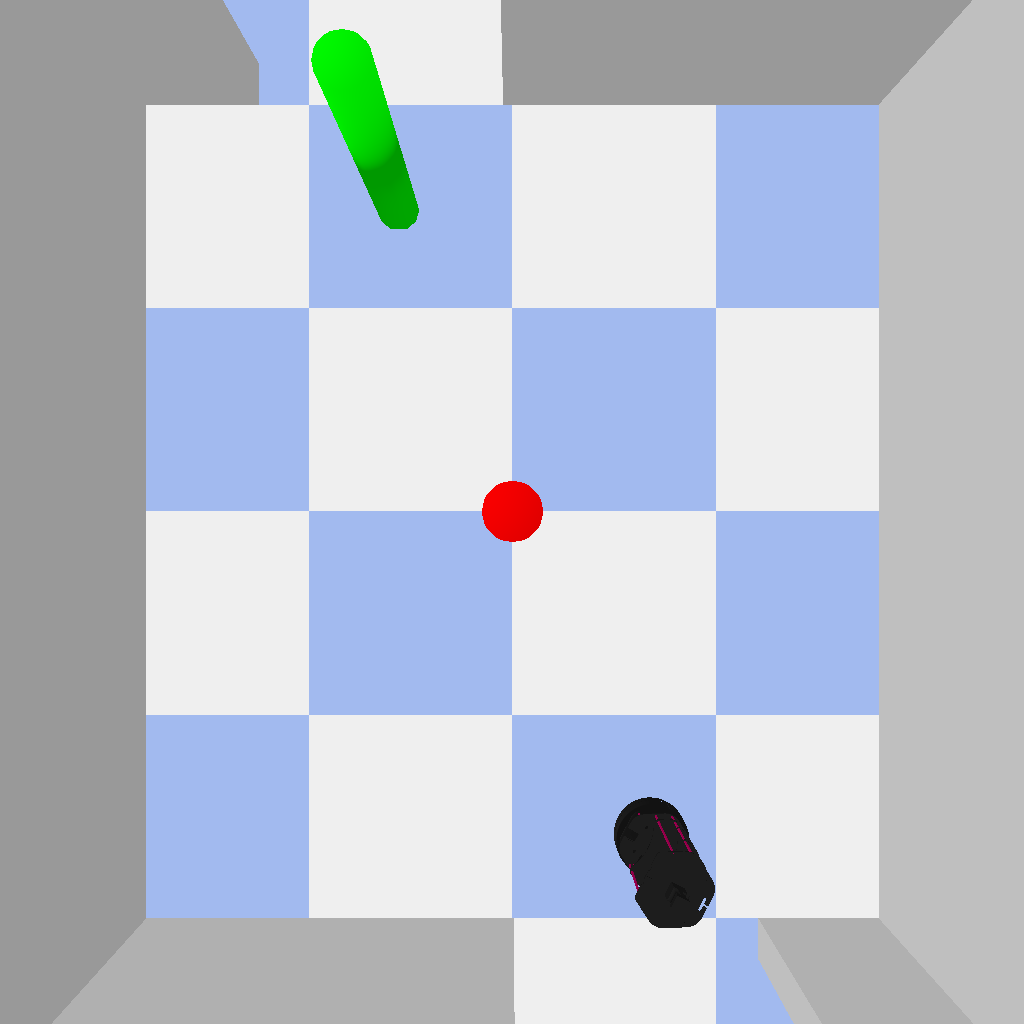}}
	\caption{Top view of both demonstration environment configurations: Corridor \textbf{(left)} and room \textbf{(right)} of the VR interface for the user study. The human needs to be avoided by the robot navigating to the goal.}
	\label{fig:environments_both} 
\end{figure}
We propose a novel VR demonstration setup, where the user teaches the robot personal navigation preferences in a virtual reality environment, see Fig.~\ref{fig:architecture}a. The user can see the robot and its navigation goal (green cone). Intuitively, the person uses the handheld controller emitting a beam of light to draw preferred trajectories onto the floor in VR. The trigger on the backside of the controller allows the user to dynamically select the robot speed along the drawn trajectory. The robot executes the demonstrated trajectory right away for reevaluation, allowing the user to either keep or redo it. After the demonstrations have been collected, the training process begins. Finally, the personalized navigation controller is evaluated in VR, before being transferred to the real robot. For the user study conducted, we chose a corridor and a room environment, see Fig.~\ref{fig:environments_both}.

\subsection{Simulator and Robot}
\label{sec:environment-robot}
Our robotic platform is the Kobuki \textit{Turtlebot 2}. As a VR and physics simulator we use Pybullet \cite{coumans_pybullet_2016}, the VR system is a HTC Vive Pro Eye.

A key challenge in using demonstrations for reinforcement learning is bridging the gap between the agent's and the demonstrator's state space. 
To do so, we analytically calculate action commands along a demonstration trajectory, so that the robot follows the trajectory by executing successive actions calculated at the control frequency $f$. 
The kinematics of a differential wheeled robot are
\begin{align}
	v &= \frac{K}{2} \left( u_r + u_l \right) \nonumber\\
	\omega &= \frac{K}{L} \left( u_r - u_l \right) \text{,} \label{eq:kinematics}
\end{align}
where $K$ is the wheel radius, $L$ the distance between both wheels, and $v$ the forward velocity. The rotation speeds of the left and right wheel are $u_l$ and $u_r$. By integrating $v$ and $\omega$ over time~$t$, we find a relation for the finite distance ${\Delta d = v \Delta t}$ traveled forward and the change in robot orientation ${\Delta \alpha= \omega \Delta t}$ within a certain time period $\Delta t$:
\begin{align}
	\label{eq:wheel_kinematics_discrete}
	\frac{v}{\omega} = \frac{\Delta d}{\Delta \alpha}
\end{align}
The time period $\Delta t$ is determined by our chosen control frequency ${f = \frac{1}{\Delta t} = \SI{5}{\hertz}}$ of the robot. Now, given a desired forward velocity $v$, one can analytically calculate the matching angular control command $\omega$ to follow a discrete segment ${(\Delta d, \Delta \alpha)}$ along a trajectory.

\subsection{Collecting and Processing Demonstration Trajectories}
\label{sec:demonstration}
% DEMO --> RL STATES, ROBOT KINEMATICS
We use the following steps to process raw demonstration trajectories into state-action pairs contained in the demonstration buffer: 
\begin{enumerate}
	\item In VR, a user draws a trajectory using the handheld controller. The analogue trigger on the controller backside allows to control the robot speed linearly in the range from $v_\text{min} =\SI{0.1}{\meter\per\second}$ to $ v_\text{max}= \SI{0.25}{\meter\per\second}$ at the drawing location. 
	\item The drawn trajectory is interpolated and smoothed with a 2D spline, parameterized by $k \in [0, 1]$. Also, the speed information is spline-interpolated. 
	\item The robot is supposed to follow the demonstrated trajectory. Based on the speed along the spline $v(k)$, we consecutively extract the locations on the spline at which the robot receives a new control command, using $\Delta d = v(k) \Delta t$.
	\item Inserting $v(k)$ for all control command locations into \eqref{eq:wheel_kinematics_discrete}, the corresponding angular velocities $\omega$ are calculated.
	\item The robot is placed and oriented according to the trajectory's starting point. 
	\item Successively, the control command tuples $a_t = (v_t, \omega_t)$ are executed and the robot follows the trajectory.
	\item Before and after the execution of each action $a_t$, we record the corresponding states $s_t$, $s_{t+1}$ and the reward $r_{t+1}$. 
	\item Finally, all state-action-reward~pairs~$\left( s_t, a_t, r_t, s_{t+1}\right)$ are stored in the demonstration buffer.
\end{enumerate}
Each demonstration trajectory is checked against possible collisions with the environment.

\subsection{Data Augmentation}
We use data augmentation to increase the data output from a single demonstration trajectory. More specifically, the robot's initial placement is shifted linearly by $\frac{\Delta d}{N_\text{aug}}$ within the distance $\Delta d = v(k_0) \Delta t$ along the spline \mbox{$N_\text{aug} = 15$ times}, where $k_0$ refers to the trajectory spline start. The result is a slightly shifted execution of the trajectory, while the original characteristic of the trajectory is preserved (${\max(\Delta d) = \SI{5}{\centi\meter} \ll }$ environment scale). Steps 5) to 8) are repeated for each data augmentation.

%\begin{figure*}[!t]
%	\centering
%	\includegraphics[width=0.95\linewidth]{figures/training}
%	\caption{Training performance distribution over all users and environments: \textbf{a)} The episode return $R$ during evaluation, \textbf{b)} episode steps during evaluation and \textbf{c)} behavior cloning loss $\mathcal{L}_\text{BC}$ during training (compare \eqref{eq:bc_loss}) are shown. We show the median and inter-quartile distance (Q1 - Q3) smoothed over 10 epochs.
%		%, where $\num{1} \text{ epoch} = \num{5000} \text{ environemnt interactions} = \num{1000} \text{ training steps}$.
%	}
%	\label{fig:training}
%\end{figure*}

\begin{figure*}[!t]
	\centering
	\includegraphics[width=0.9\linewidth]{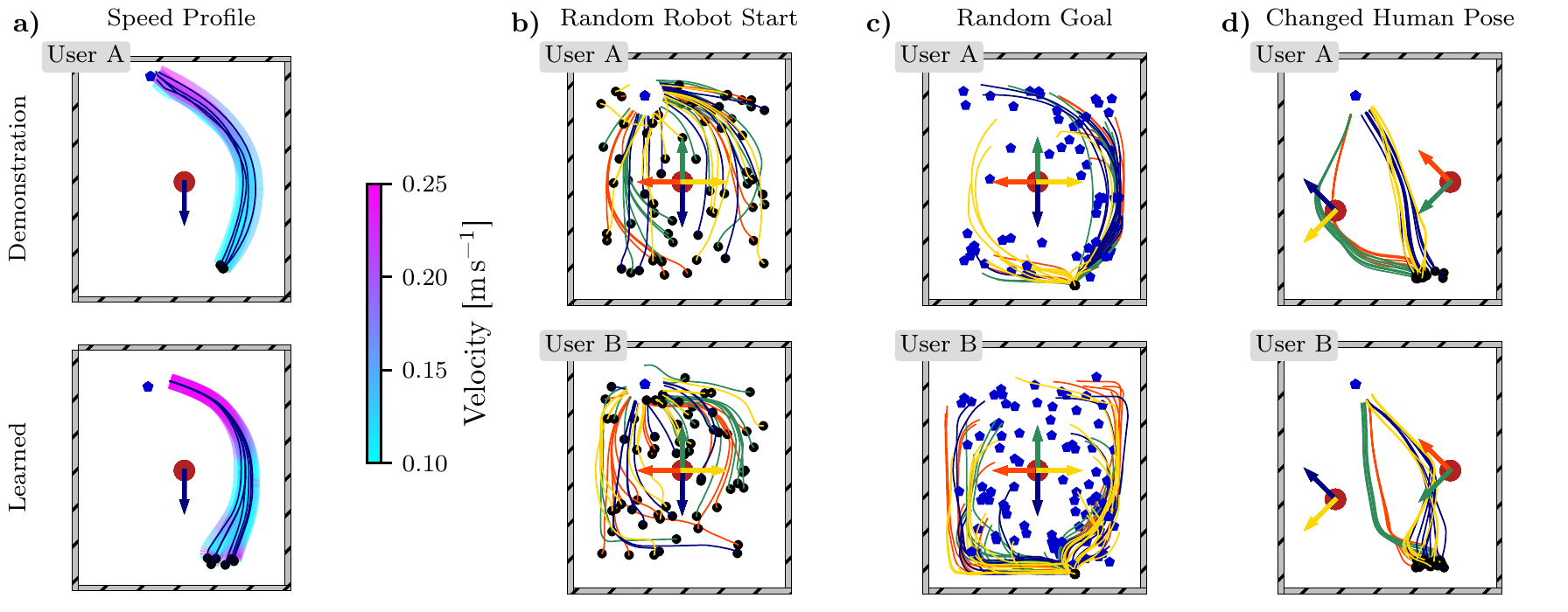}
	\caption{
		\textbf{a)} User A demonstrated a distinct speed profile \textbf{(top)} when facing the robot start position in the room environment. 
		It was successfully adapted by the learned controller \textbf{(bottom)}. 
		Furthermore, we tested the ability for generalization of the learned controller threefold by showcasing state configurations not covered by the demonstration data: 
		\textbf{b)} When the robot starts at a random position in the environment, its navigation behavior  still reflects the characteristics of the trajectory from the user demonstrations (cf. Fig.~\ref{fig:approaches}a). 
		\textbf{c)} Even when its goal is randomly placed in the room, the robot exhibits the distinct user preferences.
		\textbf{d)} The user's position and orientation  was altered to non-demonstration configurations. When the human is obstructing the robot's path while facing the wall, the robot traverses on a straight path behind the human. In all other cases, a distinct distance is kept to the human, as demonstrated by both users.
		This shows nicely how the navigation agent improved beyond the limits of the demonstration data provided.
		For a legend, please refer to Fig.~\ref{fig:approaches}.
		\label{fig:generalization}
	}
	\label{fig:training}
\end{figure*}

\subsection{Successful Demonstrations}
Reinforcement learning with demonstrations works best when demonstrations are successful, i.e., lead to the goal state. Thus, we end each demonstration trajectory with the goal state and thus a positive reward. Even if the goal position is not at the exact end of the trajectory, the goal is retroactively moved to the end of the demonstration trajectory. 
%As a result, the critic network learns to highly value states on trajectories because the demo data indicates that following the demo state-action pairs always leads to a goal state with high reward. 

\subsection{Value of Demonstration Data}
\label{sec:demo_reward}
To boost the value of demonstration-like behavior for the critics during learning, we exclusively provide the goal-reaching reward on the demonstration part of the batch, see \eqref{eq:reward_goal}.
The motivation behind this is that the agent should navigate on states $s_t$ that are as similar as possible to the states of the demonstrated trajectories, ideally recovering to those whenever useful. 
To maximize return, however, the agent generally tries to navigate towards the goal with as few state transitions as possible (due to the discount value~$\gamma$), possibly disregarding demonstrated user preferences. 
The resulting behavior corresponds to shortest-path trajectories with maximum speed while barely avoiding the human, promising a faster and higher return $R$.
The demonstration state value boost counteracts this unwanted effect, since the agent is encouraged to follow state transitions from the demonstration data due to their always \textit{higher} return.

\subsection{Training}

\label{sec:training}
% Randomization of env
We initialize the robot, human, and goal position either around the position from the demonstration configuration with probability $p_\text{env}$ or randomly in the environment to explore the entire state space with probability $(1-p_\text{env})$. 
%The motivation is that the agent learns to handle most of the state space, more precisely the entire environment. 
%The probability~$p_\text{env}$ controls the balance of two placement method: 
%i) Around the initial, predefined positions from the user demonstration with probability $(1-p_\text{env})$, or ii) randomly in the environment with probability $p_\text{env}$. 
%The same rules apply for the randomization of the static human position and orientation during training, as well as the goal position. \todo{Verify with Generalization figure}
% curriculum learning
Training starts with pre-initialization of the experience buffer by executing $\num{5e+4}$ randomly sampled actions.
Subsequently, we train for 800~epochs. 
Each epoch consists of 5000 environment interactions, while the actor and critic networks are updated every 5 interactions. 
Each epoch ends with 10 evaluation episodes.
An episode denotes the trajectory roll-out from initial robot placement until one of the end criteria is satisfied.
We train for each user and environment individually to learn context sensitive controllers. 
%The training performance in both environments is visualized in Fig.~\ref{fig:training}. 
%After training, a well performing controller is chosen based on the epoch return $R_\text{eval} = 0$, generally the last one.

An overview on all relevant and experimentally obtained training parameters can be found in \tabref{tab:training_settings}. 
We found it beneficial for the training performance to adjust the balancing factors to $\lambda_{\text{RL}}= \lambda_{\text{BC}} = \num{0.5e+1}$ at epoch 350, and reduce the actors learning rate to $l_a = \num{1e-5}$ at epoch 650.

%%%%%%%%%%%%%%%%%%%%%%%%%%%%%%%%%%%%%%%%%%%%%%%%%%%%%%%%%%%%%%%%%%%%%
% TABLES
%%%%%%%%%%%%%%%%%%%%%%%%%%%%%%%%%%%%%%%%%%%%%%%%%%%%%%%%%%%%%%%%%%%%%
\begin{table}[!b]
  \centering
  	\caption{Notations and Training settings. \label{tab:training_settings}}
	\begin{tabularx}{\linewidth}{llX}
		Notation & Value & Description \\
		\hline 
		$p_\text{env}$ & 0.25 & Placement probability: room vs. start position\\
		$n_\text{ep}$ & 300 & Maximum number of steps per episode\\
		$B_{E}$ & \num{1e+6} & Experience replay buffer size\\
		%$B_{D}$ & & Demonstration buffer size\\
		% $b_{E/D}$ & 64 & Batch size of experience/demo data\\
		$l_a$ & \num{1e-4}  & Learning rate of actor\\
		$l_c$ & \num{8e-4}  & Learning rate of critic\\
		$\gamma$ & 0.99 & Discount factor\\
		$\sigma_{\epsilon_\pi}$ & 0.2 & Std. deviation of exploration noise $\epsilon_\pi$ \\
		$\sigma_{\epsilon_\theta}$ & 0.05 & Std. deviation of target policy noise $\epsilon_\theta$ \\
		% $\tau$ & 0.005 & Target slow-moving update rate\\
		% $c_\text{rew}$ & 5 & Reward base value\\
		$\lambda_{\text{RL}}$ & 10/3 & Weighting factor of RL gradient on actor\\
		$\lambda_{\text{BC}}$ & 20/3 & Weighting factor of BC loss gradient on actor\\		
		\hline
	\end{tabularx}
\end{table}
%%%%%%%%%%%%%%%%%%%%%%%%%%%%%%%%%%%%%%%%%%%%%%%%%%%%%%%%%%%%%%%%%%%%%

\section{Experimental Evaluation}
\label{sec:exp}
\begin{figure}[t]
	\centering
	\includegraphics[width=0.95\linewidth]{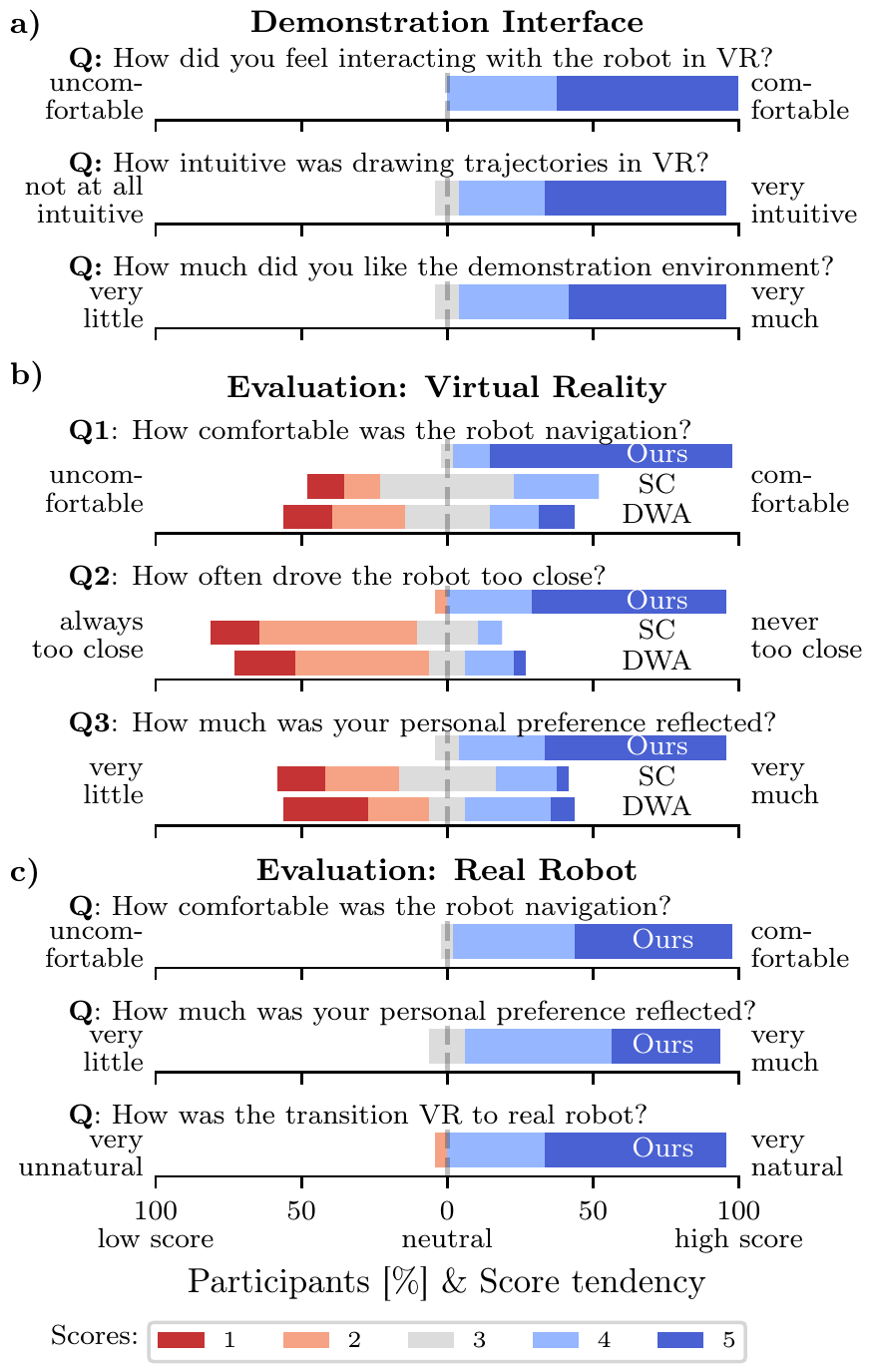}
	\caption{User study survey results of both the demonstration and evaluation session. \textbf{a)} The demonstration interface was predominantly appreciated and experienced intuitive by the participants. \textbf{b)} Evaluation: In virtual reality, both the Dynamic Window Approach and the social cost model were outperformed by our personalized controller in various aspects. \textbf{c)} On the real robot, our novel personalized controller was perceived predominantly positive as well. The plot bar's positions are aligned to the neutral score~(3) to indicate overall rating.
	\label{fig:survey_all}}
\end{figure}
This section highlights the results of our user study  and provides a qualitative and quantitative analysis of the learned personalized navigation controller.
% USER STUDY
\subsection{User Study}
We conducted a user study with 24 non-expert participants~(13 male, 11 female) to i) record individual navigation preferences (demonstration data), ii) evaluate the navigation behavior learned by our personalized controller, and iii) evaluate the presented VR demonstration interface.
Participants attended two appointments, the first being the demonstration session and the second being the evaluation session. 
In the user study section, the values in brackets refer to the mean survey-scores (1-5) and their standard deviation.

% Recording sesseion
\subsubsection{Demonstration Session}
During the demonstration session, trajectories in both environments (corridor and room) were recorded, see Fig.~\ref{fig:approaches}a. 
Each environment featured four position-orientation pairs (color-coded) for the participant.
For each pair, between three and five trajectories were recorded. The total time investment was about 20 min.
% The trajectories' color-shading indicates the speed $v$ of the robot at a given position, where blue is slow and violet fast.
After the recording session, the participants were asked about their experience with the VR demonstration interface. 
The survey questions and results are shown in Fig.~\ref{fig:survey_all}a).
Participants predominantly experienced comfortable interactions with the simulated robot ($\num{4.6 \pm 0.1}$) and found drawing trajectories with our interface very intuitive ($\num{4.5 \pm 0.1}$).
Also, no participants disliked our demonstration environments while the majority liked it \textit{very much} ($\num{4.6 \pm 0.1}$).  

% Eval sesseion VR
\subsubsection{Evaluation Session}
During the second session, our personalized navigation approach was evaluated against two approaches in virtual reality: The Dynamic Window Approach (DWA) \cite{fox_dynamic_1997} using the ROS \textit{move\_base} package \cite{quigley_ros_2009} in combination with a 2D lidar sensor, and a social cost model~(SC) based on the configuration of~\cite{kollmitz_time_2015}.
Each navigation approach was shown in VR (order: SC $\rightarrow$ DWA $\rightarrow$ Ours) in both environments for all four position-orientation pairs (cf. Fig.~\ref{fig:approaches}b-d), followed by an evaluation survey (cf. Fig.~\ref{fig:survey_all}b). Potential ordering effects cannot be completely ruled out. Participants were unaware of presented approach types. 
Pairwise Bonferroni-corrected Wilcoxon signed-rank tests indicated that our personalized approached significantly outperformed both the SC and DWA navigation on all three measures comfort (Q1), unpleasant closeness (Q2) and preference (Q3) (see \tabref{tab:survey_statistics}). 
%Whereas the participants perceived the robot to navigate often too close during DWA and SC (), they hardly ever judged it too close with our personalized approach (). 
No significant differences were measured between SC and DWA.

%Concerning the distance the robot kept to the participants during navigation, both the SC and DWA navigation were perceived as too close (\num{2.2 \pm 0.2}$ and $\num{2.4(2)}$). In contrast, the robot hardly ever navigated to close with our personalized approach $\num{4.6 \pm 0.2}$. No significant difference in closeness perception was measured between SC and DWA.
%Finally, the participants rated whether their preference was reflected in the presented approaches. 
%Here, the results are in favor of our personalized controller with an averaged score increase of $\num{1.9 \pm 0.4}$ compared to SC and DWA.

% Eval sesseion RR
\subsubsection{Real Robot Evaluation}
Our personalized controller was demonstrated on the real robot (room environment) to investigate the participants transition experience from the simulated to the real robot. 
% Here, the robot's odometry is obtained from the VR tracking system using a robot-mounted generic tracking device.
The real robot evaluation was also complemented by a survey, see Fig.~\ref{fig:survey_all}c). As in~VR, the navigation of the real robot  was predominantly experienced comfortable ($\num{4.5 \pm 0.1}$) and participants saw their preferences mostly reflected ($\num{4.3 \pm 0.1}$). Furthermore, the transition from the simulated robot experience in~VR to the real robot was mostly experienced as \textit{very natural} ($\num{4.5 \pm 0.1}$).

%%%%%%%%%%%%%%%%%%%%%%%%%%%%%%%%%%%%%%%%%%%%%%%%%%%%%%%%%%%%%%%%%%%%%
% TABLES
%%%%%%%%%%%%%%%%%%%%%%%%%%%%%%%%%%%%%%%%%%%%%%%%%%%%%%%%%%%%%%%%%%%%%
\begin{table}[!b]
	\centering
	\caption{Wilcoxon signed-rank tests on mean scores of all approaches \label{tab:survey_statistics}}
	\begin{tabularx}{\linewidth}{XXXX}
		Question & Ours - SC & Ours - DWA & SC - DWA\\
		\hline 
		Q1: comfort & $z = \num{-4.17}^*$ & $z = \num{-4.01}^*$ & $z = \num{-1.81}$ \\	
		Q2: closeness & $z = \num{-4.29}^*$ & $z = \num{-4.2}^*$ & $z = \num{-3.61}$ \\	
		Q3: preference & $z = \num{-4.01}^*$ & $z = \num{-4.06}^*$ & $z = \num{-1.97}$ \\	
		\hline
		\multicolumn{4}{>{\hsize=\dimexpr4\hsize+4\tabcolsep+\arrayrulewidth\relax}X}{Note that statistical significance was always $p<0.001$, as marked with *. All other comparisons did not reach statistical significance.}
	\end{tabularx}
\end{table}
%%%%%%%%%%%%%%%%%%%%%%%%%%%%%%%%%%%%%%%%%%%%%%%%%%%%%%%%%%%%%%%%%%%%%

\subsection{Qualitative Navigation Analysis}
% Preference description
Fig.~\ref{fig:approaches}a shows demonstration data from two participants in both environments. 
In the room environment, the preference of participant A is a smooth curve around their position, while the robot drives in their field of view when approaching from either side. 
%In the corridor, a distinct speed profile accelerating the robot when it has passed the participant's position was demonstrated.
Interestingly, participant B's preference is a wall-following robot that navigates at higher distance to the human, compared to participant A.

% Preference reproduction 
Fig.~\ref{fig:approaches}b shows trajectories of the learned navigation behavior. The learned policy clearly reflects the characteristics of the demonstration trajectories. Furthermore, the robot adjusts its navigation trajectory according to the human orientation. For user A, it learned to traverse in the user's field of view, compare yellow orientation and trajectories. In participant B's demonstration, trajectories from a single position-orientation pair traverse both in front and behind the participant. Here, no specific side preference is given and the controller reproduces trajectories mainly on one side.

% Speed
Beside trajectory shape, user demonstrated speed profiles along the demonstration trajectories. As an example, Fig.~\ref{fig:generalization}a) depicts how user A demonstrated a distinct speed profile when directly facing the robot start position in the room environment. After the robot slowly approached and passed by, it was allowed to accelerate. As can be seen, the behavior is picked up by the controller during training.

% Plots with barh
\subsection{Quantitative Navigation Analysis}
Fig.~\ref{fig:approaches}e-g compare quantitative properties of all three evaluation approaches and demonstrations from all 24 study participants. 
%We distinguish between both environments. 
The personalized navigation trajectories are on average longer than those by DWA or SC, while maintaining a higher minimal distance to the human.
Interestingly, the mean preferred minimum human distance gathered from the user demonstrations is similar in both environments, averaged at $\overline{d_H} = \SI{1.1 \pm 0.2}{\meter}$.
The path area is calculated between the trajectory and linear distance from start to goal. A higher path area reveals earlier deviation from the linear path in favor of personalization, as it is the case for our personalized controller, compare Fig.~\ref{fig:approaches}g. This clearly indicates that users prefer personalized navigation trajectories over shortest path navigation. Furthermore, the large standard deviation of the path area indicates a high trajectory shape variability among the participants.

% Generalization to unseen states and constellations
%\begin{figure}[t]
%	\centering
%	\includegraphics[width=0.9\linewidth]{figures/generalization}
%	\caption{Generalization of navigation behavior, when the robot is placed randomly in the environment (simulated). The navigation agent in general still reflects the user preference demonstrations (compare Fig.~\ref{fig:approaches}a) by approaching demonstration-like states, i.e., trajectories. When appropriate, it drives straight to the goal. This shows nicely how the navigation agent improved beyond the limits of such little demonstration data, thus the successful transfer from few user-preferences to a robust navigation controller,
%		\label{fig:generalization}}
%\end{figure}

\subsection{Generalization}
Finally, we tested the ability for generalization of the learned navigation policy, see Fig.~\ref{fig:generalization}b-d. First, the robot started at random positions in the environment not covered by the demonstrations. As can be seen, the controller still reflects the user preferences in the driving style (cf. Fig.~\ref{fig:generalization}b and Fig.~\ref{fig:approaches}a) by either approaching demonstration-like states, or reproducing demonstration-like navigation patterns at slightly different positions in the environment. When appropriate, the robot drives straight to the goal. 
Second, we tested random goal positions in the environment~(cf. Fig.~\ref{fig:generalization}c). Interestingly, only after driving in accordance with preferences, the robot turns towards the goal when in direct vicinity.
Finally, we tested altered human positions~(cf. Fig.~\ref{fig:generalization}d). Human position-orientation pairs not covered in the demonstration data encourage the controller to still keep a preference-like distance.

As demonstrated with these results, our framework can successfully learn a personalized navigation controller that improves beyond the limits of few demonstration trajectories.

%%%%%%%%%%%%%%%%%%%%%%%%%%%%%%%%%%%%%%%%%%%%%%%%%%%%%%%%%%%%%%%%%%%%%%%%%%%%%%%%
\section{Conclusion}
\label{sec:conclusion}
To summarize, we presented both a learning framework and an intuitive virtual reality interface to teach navigation preferences to a mobile robot. 
From a few demonstration trajectories, our context-based navigation controller successfully learns to reflect user-preferences and furthermore transfers smoothly to a real robot. 
The conducted user study provides evidence that our personalized approach significantly surpasses standard navigation approaches in terms of perceived comfort. Furthermore, the study verifies the demand for personalized robot navigation among the participants.
Our results are a first important step towards personalized robot navigation, made possible by our interface and user study. As a next logical step, we will transfer the framework to more complex and diverse environments.  

%%%%%%%%%%%%%%%%%%%%%%%%%%%%%%%%%%%%%%%%%%%%%%%%%%%%%%%%%%%%%%%%%%%%%%%%%%%%%%%%
% Only if applicable
% \section*{Acknowledgments}

\bibliographystyle{IEEEtran}
\bibliography{bibliography}

\end{document}